\documentclass[runningheads]{llncs}

\usepackage{eccv}

\usepackage{eccvabbrv}

\usepackage{graphicx}
\usepackage{booktabs}

\usepackage[accsupp]{axessibility}  % Improves PDF readability for those with disabilities.

\usepackage{hyperref}

\usepackage{orcidlink}

\usepackage{multirow}

\begin{document}

% ---------------------------------------------------------------
% TODO REVIEW: Replace with your title
\title{AEP$n$P: A Less-constrained EP$n$P Solver for Pose Estimation with Anisotropic Scaling} 

% TODO REVIEW: If the paper title is too long for the running head, you can set
% an abbreviated paper title here. If not, comment out.
\titlerunning{AEP$n$P}

% TODO FINAL: Replace with your author list. 
% Include the authors' OCRID for the camera-ready version, if at all possible.
\author{Jiaxin Wei\inst{1}\orcidlink{0000-0002-7645-6093} \and
Stefan Leutenegger\inst{1}\orcidlink{0000-0002-7998-3737} \and
Laurent Kneip\inst{2}\orcidlink{0000-0001-6727-6608}}

% TODO FINAL: Replace with an abbreviated list of authors.
\authorrunning{J.~Wei et al.}
% First names are abbreviated in the running head.
% If there are more than two authors, 'et al.' is used.

% TODO FINAL: Replace with your institution list.
\institute{Smart Robotics Lab, CIT, Technical University of Munich \\
\email{\{jiaxin.wei, stefan.leutenegger\}@tum.de} \and
Mobile Perception Lab, SIST, ShanghaiTech University \\
\email{lkneip@shanghaitech.edu.cn}}

\maketitle

\begin{abstract}
  Perspective-$n$-Point (P$n$P) stands as a fundamental algorithm for pose estimation in various applications. In this paper, we present a new approach to the P$n$P problem with relaxed constraints, eliminating the need for precise 3D coordinates, which is especially suitable for object pose estimation where corresponding object models may not be available in practice. Built upon the classical EP$n$P solver, we refer to it as AEP$n$P due to its ability to handle unknown anisotropic scaling factors in addition to the common 6D transformation. Through a few algebraic manipulations and a well-chosen frame of reference, this new problem can be boiled down to a simple linear null-space problem followed by point registration-based identification of a similarity transformation. Experimental results on both simulated and real datasets demonstrate the effectiveness of AEP$n$P as a flexible and practical solution to object pose estimation. Code: \url{https://github.com/goldoak/AEPnP}.
  \keywords{Perspective-$n$-Point \and Object pose estimation}
\end{abstract}

\section{Introduction}
\label{sec:intro}

Camera pose estimation is an active area of research in both the computer vision and robotics communities as it is indispensable in many practical applications such as visual localization, 3D reconstruction, and object tracking. The Perspective-$n$-Point problem is one of the most well-known techniques in this field, utilizing 2D image measurements and corresponding 3D landmarks in the environment to infer the camera pose. Many variants have been introduced over the years to continuously improve accuracy, efficiency, and robustness~\cite{lepetit09,hesch11,kneip2014upnp}. 

In recent instance-level object pose estimation, a main underlying assumption is that a 3D model exists for each observed object such that the problem can be easily solved by traditional pose solvers. However, this imposes significant limitations on the applicability of those instance-level methods, especially in scenarios where obtaining an exact 3D model is challenging. Given a fully calibrated camera, traditional P$n$P solvers require at least three or more 2D-3D correspondences for accurate pose estimation. The absence of precise 3D coordinates can lead to sub-optimal results or even failure of existing P$n$P solvers. 

\begin{figure}[htbp]
  \centering
   \includegraphics[width=\linewidth]{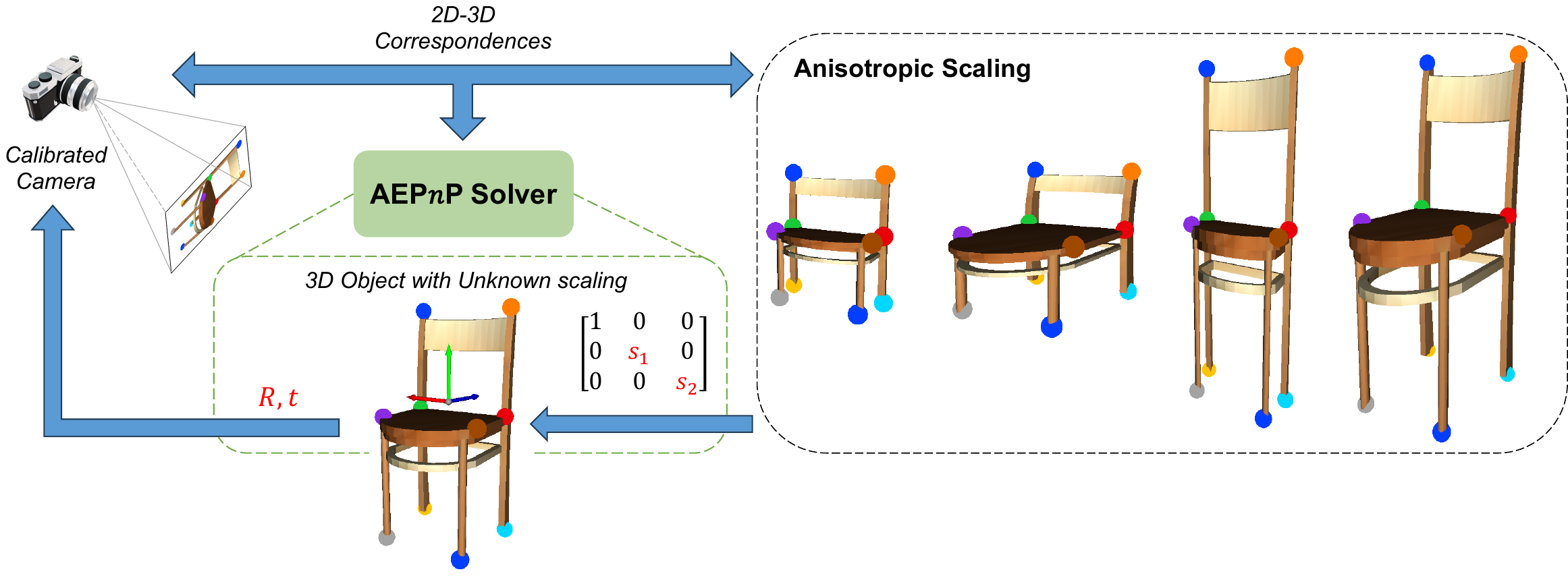}
   \caption{Illustration of one practical case in object pose estimation which can be well handled by our proposed AEP$n$P solver. Specifically, it can simultaneously estimate anisotropic scaling factors as well as the 6D pose.}
   \label{Fig:front_fig}
\end{figure}

To relax this strict assumption, we propose a novel and less-constrained solver, \textit{AEP$n$P} (Anisotropic EP$n$P), allowing for more flexible estimation of 6D pose and potential anisotropic scaling lying in 2D-3D correspondences. As outlined in Fig.\ref{Fig:front_fig}, the exact 3D model of an object may not be available in some practical scenarios (e.g.\ CAD model retrieval), but only a 3D shape prior is given up to unknown aspect ratios. AEP$n$P addresses this case by simultaneously estimating the scaling factors along different axes such that the 3D shape prior can be aligned to the observed object in the 2D image after non-uniform stretching or compressing. In the following section, we will demonstrate that with just a few manipulations, the equations for the above problem can be brought into a very similar form to the well-known EP$n$P algorithm~\cite{lepetit09}, and therefore simply solved by a linear null-space decomposition.

We conduct thorough experiments on simulation data to analyze the impact of noise and the number of correspondences used for estimation, clearly demonstrating the superiority of AEP$n$P over EP$n$P without compromising computational speed when precise 2D-3D correspondences are unavailable. We further integrate the AEP$n$P solver into a RANSAC loop to handle outlier correspondences and prove its effectiveness on two real datasets. To establish its practical applicability, we also design a challenging experiment, leveraging only sparse keypoints to perform object pose estimation with unknown anisotropic scaling. In summary, we make the following contributions:
\begin{enumerate}
    \item We relax the constraints imposed by traditional P$n$P solvers and propose a novel AEP$n$P algorithm that can handle unknown anisotropic scaling while estimating the camera pose, providing a more flexible solution to object pose estimation task.
    \item Through algebraic manipulations, the AEP$n$P solver establishes a highly similar form to the well-known EP$n$P algorithm and then can be solved by a simple linear null-space decomposition followed by scale-aware control point registration.
    \item Comprehensive experiments on both simulation and real data prove the robustness and effectiveness of our AEP$n$P, showcasing its valuable advancement in the field of object pose estimation.
\end{enumerate}

\section{Related Work}
\label{sec:relatedWork}

Camera pose estimation is first introduced within the photogrammetry community~\cite{grunert1841,finsterwalder37,merrit49}. These early solutions including the first minimal solvers developed within the computer vision community~\cite{fischler81,hung86,linnainmaa88,grafarend89} are well reviewed in the work of Haralick et al.~\cite{haralick91}. De~Menthon and Davis~\cite{dementhon92} further propose exact and approximate solutions to the so-called Perspective-three-point (P3P) problem, while Gao et al.~\cite{gao03} present the first complete solution classification, analyzing the number of solutions as a function of both algebraic and geometric conditioning. Kneip et al.~\cite{kneip13} contribute the first solution to the P3P problem that directly solves the camera pose parameters rather than the traditional formulation in terms of unknown depths along rays. The P3P problem remains one of the most fundamental geometric computer vision problems til modern times, as novel formulations that improve on execution speed, numerical robustness, or explainability of the solution number keep being discovered~\cite{ke17,banno18,persson18,nakano19,ding23}.

The P3P problem is a special case of the P$n$P problem. The first solution for P$n$P is the popular DLT algorithm introduced by Sutherland in 1963~\cite{sutherland63}. Improved solutions taking into account known camera calibration parameters are proposed later by Dhome et al.~\cite{dhome89}, De~Menthon and Davis~\cite{dementhon95}, and Horaud et al.~\cite{horaud89}. The linear solutions to the problem are further investigated by Quan and Lan~\cite{quan99} and Ameller et al.~\cite{ameller00}, Fiore~\cite{fiore01}, and Ansar and Daniilidis~\cite{ansar03}, while Schweighofer and Pinz~\cite{schweighofer06} solve the P$n$P problem based on a geometric error criterion. Notably, Lepetit et al.~\cite{lepetit09} propose another linear complexity solution to the P$n$P problem which is still considered as state-of-the-art when efficiency is the major concern. This solver also serves as the main starting point for the method proposed in this paper. Another breed of solvers is later introduced by Hesch and Roumeliotis~\cite{hesch11}, Zheng et al.~\cite{zheng13}, and Kneip et al.~\cite{kneip2014upnp}. These solvers have linear complexity and find the global minimum of the sum of squared object space errors in closed form. 

In this work, we consider a rather new problem that takes into account two unknown anisotropic scales of the reference model along the $y$ and $z$-axis, which has certain similarities with camera pose estimation algorithms that address the partially calibrated case. Abidi and Chandra~\cite{abidi95} solve the problem of camera pose and unknown focal length for planar point distributions while the first solution for non-planar scenes is introduced by Triggs~\cite{triggs99}. Later, Bujnak et al.~\cite{bujnak08}, Zheng et al.~\cite{zheng14}, and Wu~\cite{wu15} propose the first minimal solutions that work in both planar and non-planar scenarios. Several non-minimal solutions continue dealing with only one unknown effective focal length (i.e. assume the camera has square pixels)~\cite{penate13,kanaeva15,zheng16}. Back in 1999, Triggs~\cite{triggs99} has already pointed out a possible solution for handling aspect ratio. However, the exact constraints that are required to solve for the pose, focal length, and aspect ratio are introduced much later by Guo~\cite{guo2013novel}. Larsson et al. \cite{larsson2018camera} consider the case where focal length and aspect ratio are estimated alongside the principal point. To solve this problem, they construct a minimal solver using 5-point correspondences.

\section{Theory}
\label{sec:theory}

We start by reviewing the classical EP$n$P algorithm and introduce all relevant notations. Then, we extend the basic problem formulation to accommodate the anisotropic scaling case. Finally, the 6D pose and scaling factors are derived through a few simple algebraical manipulations.

\subsection{Problem Formulation and Review of EP$n$P}

Consider a fully calibrated camera $\mathcal{C}$ and a set of 2D-3D correspondences $\{\mathbf{u}_i \leftrightarrow \mathbf{x}_i\}$, where $\mathbf{u}_i$ are normalized image coordinates obtained through the known camera intrinsics. The basic relationship is then given by:
\begin{equation}
\lambda_i \mathbf{u}_i = \mathbf{R} \mathbf{x}_i + \mathbf{t},
\label{eq:firstForm}
\end{equation}
where $\lambda_i$ represents the unknown depth of the world point $\mathbf{x}_i$ as seen from the camera frame. The rotation $\mathbf{R}$ and translation $\mathbf{t}$ transform points from the world frame to the camera frame. EP$n$P simplifies this by introducing a low-dimensional linear parametrization. Each world point $\mathbf{x}_i$ is expressed as a linear combination of four control points $\mathbf{c}_j$:
\begin{equation}
\mathbf{x}_i  =  \sum_{j=0} ^ 3 \alpha_{ij} \mathbf{c}_j
\end{equation}
where the coefficients $\alpha_{ij}$ satisfy:
\begin{equation}
\sum_{j=0}^3 \alpha_{ij} = 1.
\end{equation}
One common way to choose the control points is decomposing each world point into a centroid $\mathbf{c}_0$ and a linear combination of basis components $\sigma_j$. Specifically, we have
$\mathbf{x}_i = \mathbf{c}_0 + \sum_{j=1}^3 \alpha_{ij}\sigma_j$
where each basis component $\sigma_j$ can be written as the difference between a control point and the centroid as
$\sigma_j = \mathbf{c}_j - \mathbf{c}_0.$

Now we reformulate the incidence relation \eqref{eq:firstForm} as a function of the control points in the camera frame without affecting the linear combination weights:
\begin{eqnarray}
  \lambda_i \mathbf{u}_i  =  \mathbf{R}\mathbf{x}_i + \mathbf{t} 
  & = & \mathbf{R} \left( \sum_{j=0}^3 \alpha_{ij} \mathbf{c}_j\right) + \left(\sum_{j=0}^{3} \alpha_{ij} \right) \mathbf{t} \\
  & = & \sum_{j=0}^3 \alpha_{ij} \left( \mathbf{R} \mathbf{c}_j + \mathbf{t} \right) \\
  & = & \sum_{j=0}^3 \alpha_{ij} \mathbf{c}_j^\mathcal{C}\label{eq:controlPointForm}
\end{eqnarray}
where $\mathbf{c}_j^\mathcal{C}$ are the control points in the camera frame. Once we solve for those control points in the camera frame, the camera pose can be easily derived through alignment between $\mathbf{c}_j^\mathcal{C}$ and $\mathbf{c}_j$. While we need to first eliminate the unknown depth parameters $\lambda_i$. Given that the third component of $\mathbf{u}_i$ is 1, we can derive from \eqref{eq:controlPointForm} that:
\begin{equation}
  \lambda_i = \sum_{j=0}^3 \alpha_{ij}c_{jz}^{\mathcal{C}}
\end{equation}
where $c_{jz}^{\mathcal{C}}$ represents the z-axis coordinate of $\mathbf{c}_j^\mathcal{C}$. Back substitution this into the original equation leads to
\begin{eqnarray}
  \sum_{j=0}^{3}\left(\alpha_{ij}c_{jx}^{\mathcal{C}}-u_{ix}\alpha_{ij}c_{jz}^{\mathcal{C}}\right) & = & 0\label{eq:epnp_eq1} \\
  \sum_{j=0}^{3}\left(\alpha_{ij}c_{jy}^{\mathcal{C}}-u_{iy}\alpha_{ij}c_{jz}^{\mathcal{C}}\right) & = & 0\label{eq:epnp_eq2}
\end{eqnarray}
where $u_{ix}$ and $u_{iy}$ represent the x-axis and y-axis coordinates of $\mathbf{u}_i$, respectively. For $n$ correspondences, we can set up a linear system and solve for the control points in the camera frame using linear null-space decomposition:
\begin{equation}
    \mathbf{A} \left[\begin{matrix}
        \mathbf{c}^{\mathcal{C}}_0 \\
        \mathbf{c}^{\mathcal{C}}_1 \\
        \mathbf{c}^{\mathcal{C}}_2 \\
        \mathbf{c}^{\mathcal{C}}_3
    \end{matrix}\right] = \mathbf{0}
\end{equation}
where $\mathbf{A}$ is a $2n \times 12$ matrix obtained from the coefficients of \eqref{eq:epnp_eq1} and \eqref{eq:epnp_eq2} for each correspondence.

% \begin{equation}
%     \left[\begin{matrix}
%         \alpha_{00} \mathbf{I}_2 & -\alpha_{00} \left[\begin{matrix}u_{0x} \\ u_{0y}\end{matrix}\right] & \cdots & \alpha_{03} \mathbf{I}_2 & -\alpha_{03} \left[\begin{matrix}u_{0x} \\ u_{0y}\end{matrix}\right] \\
%         & & \vdots & & \\
%         \alpha_{n0} \mathbf{I}_2 & -\alpha_{n0} \left[\begin{matrix}u_{nx} \\ u_{ny}\end{matrix}\right] & \cdots & \alpha_{n3} \mathbf{I}_2 & -\alpha_{n3} \left[\begin{matrix}u_{nx} \\ u_{ny}\end{matrix}\right]
%     \end{matrix}\right] \left[\begin{matrix}
%         \mathbf{c}_0 \\
%         \mathbf{c}_1 \\
%         \mathbf{c}_2 \\
%         \mathbf{c}_3
%     \end{matrix}\right] = \mathbf{0}.
% \end{equation}

% In the original work of \cite{lepetit09}, further variants of the solver are proposed for planar cases or to find solutions as linear combinations of multiple null-space vectors, enhancing robustness and accuracy under noise or near-rank deficient scenarios.

\subsection{Extension to Anisotropic Scaling Case}

In this paper, we address a generalization of the P$n$P problem that is applicable when the corresponding 3D coordinates are affected by unknown anisotropic scaling. This will introduce additional scaling parameters into the transformation between the world and the camera frame. These scaling factors act along each dimension independently. Algebraically, this modifies the original EP$n$P problem formulation \eqref{eq:firstForm} to:
\begin{equation}
\lambda_i \mathbf{u}_i = \mathbf{R} \mathbf{S} \mathbf{x}_i + \mathbf{t},
\label{eq:scaledForm}
\end{equation}
where $\mathbf{S}$ is a $3 \times 3$ diagonal matrix representing the scaling factors along each dimension. Similar to \eqref{eq:controlPointForm}, we can reformulate \eqref{eq:scaledForm} as a function of control points, which gives us:
\begin{eqnarray}
  \lambda_i \mathbf{u}_i  =  \mathbf{R}\mathbf{S}\mathbf{x}_i + \mathbf{t} 
  & = & \mathbf{R}\mathbf{S} \left( \sum_{j=0}^3 \alpha_{ij} \mathbf{c}_j\right) + \left(\sum_{j=0}^{3} \alpha_{ij} \right) \mathbf{t} \\
  & = & \sum_{j=0}^3 \alpha_{ij} \left( \mathbf{R}\mathbf{S} \mathbf{c}_j + \mathbf{t} \right) \\
  & = & \sum_{j=0}^3 \alpha_{ij} \mathbf{\hat{c}}_j^\mathcal{C}\label{eq:controlPointForm2}
\end{eqnarray}
where $\mathbf{\hat{c}}_j^\mathcal{C}$ are the transformed, rescaled control points in the camera frame. This allows us to incorporate anisotropic scaling into the estimation process, providing a more flexible and accurate solution for scenarios where scaling varies along different dimensions.

% $\mathbf{S}=\left[\begin{matrix} s_1 & 0 & 0 \\ 0 & s_2 & 0 \\ 0 & 0 & s_3 \end{matrix}\right]$

\subsection{Retrieval of 6D Pose and Scaling Parameters}

The final camera pose and scaling parameters can be determined through a scale-aware alignment between the control points $\mathbf{\hat{c}}_j^\mathcal{C}$ in the camera frame, and the control points $\mathbf{c}_j$ defined in the world frame. The relationship between these points is given by:
\begin{equation}
  \mathbf{\hat{c}}_j^{\mathcal{C}} = \mathbf{R}\mathbf{S} \mathbf{c}_j + \mathbf{t}
\end{equation}
We can easily solve the above alignment via a straightforward selection of the control points in the world frame. Assuming that the 3D coordinates are centered around $\mathbf{c}_0=\mathbf{0}$ and the other three control points are simply defined as
\begin{equation}
    \mathbf{c}_1 = \begin{bmatrix}1 & 0 & 0\end{bmatrix}^T,
    \mathbf{c}_2 = \begin{bmatrix}0 & 1 & 0\end{bmatrix}^T,
    \mathbf{c}_3 = \begin{bmatrix}0 & 0 & 1\end{bmatrix}^T.
\end{equation}
Using the rotation matrix $\mathbf{R} = \begin{bmatrix}\mathbf{r}_1 & \mathbf{r}_2 & \mathbf{r}_3\end{bmatrix}$, we can derive the following:
\begin{equation}
    \mathbf{\hat{c}}_0^{\mathcal{C}} = \mathbf{R}\mathbf{S}\cdot\mathbf{0} + \mathbf{t}  \Rightarrow  \mathbf{t} = \mathbf{\hat{c}}_0^{\mathcal{C}}
\end{equation}
and
\begin{eqnarray}
    \mathbf{\hat{c}}_j^{\mathcal{C}} = s_j \mathbf{r}_j + \mathbf{t}
    & \Rightarrow & s_j \mathbf{r}_j = \mathbf{\hat{c}}_j^{\mathcal{C}} - \mathbf{\hat{c}}_0^{\mathcal{C}} \\
    & \Rightarrow & \left\{\begin{matrix} s_j &=& \| \mathbf{\hat{c}}_j^{\mathcal{C}} - \mathbf{\hat{c}}_0^{\mathcal{C}} \| \\
    \mathbf{r}_j &=& \frac{\mathbf{\hat{c}}_j^{\mathcal{C}} - \mathbf{\hat{c}}_0^{\mathcal{C}}}{\| \mathbf{\hat{c}}_j^{\mathcal{C}} - \mathbf{\hat{c}}_0^{\mathcal{C}} \|}\end{matrix}\right. \text{when}~j=1,2,3
\end{eqnarray}
It is important to note that the solution is affected by an overall scale ambiguity, as the three scaling factors can be scaled together with the translation. For convenience, we assume that the coordinates along the $x$-axis are known without ambiguity, while the coordinates along the $y$ and $z$-axis are affected by two independent scale factors, respectively.

% Additionally, the solution can be refined by enforcing the orthonormality of the rotation matrix $\mathbf{R}$. Orthogonality constraints, such as $(\mathbf{c}_{j1}-\mathbf{c}0) \cdot (\mathbf{c}{j2}-\mathbf{c}_0) = 0$, could also be added to explore higher-dimensional null-space vector combinations. However, this refinement is not explored in the current work.

\section{Experiments}
\label{sec:experiments}

In this section, we analyze the performance of AEP$n$P on both simulated data and real datasets, and demonstrate its practical applicability in a challenging experiment using only sparse keypoints to infer object pose and unknown anisotropic scaling factors.

\paragraph{\textbf{Comparisons}}
Considering that we are dealing with a new problem arising from object pose estimation, and there are no suitable solvers available, we primarily use two methods for comparison, i.e. EP$n$P and AEP$n$P. As known, the EP$n$P method requires precise 2D-3D correspondences to function properly. Hence, we only utilize it as a rough baseline. We also equip AEP$n$P with a RANSAC loop to handle potential noise and outlier correspondences in the real datasets effectively. To improve the performance of RANSAC-AEP$n$P, we further introduce a non-linear refinement step, optimizing over desired unknown variables, including rotation, translation, and anisotropic scaling factors.

\paragraph{\textbf{Metrics}}
We employ a set of metrics to evaluate the performance of our proposed method. Specifically, we report three key metrics for all experiments:
\begin{itemize}
    \item Rotation error: $$R_{err} = \arccos(\mathrm{trace}(\frac{\mathbf{R}_{gt}^T\mathbf{R}-1}{2}))$$
    \item Translation error: $$t_{err} = \|\mathbf{t}-\mathbf{t}_{gt}\|_2$$
    \item Anisotropic $s_1/s_2$ error: $$s_{err} = \frac{|s-s_{gt}|}{s_{gt}}$$
\end{itemize}

\subsection{Simulation Data}\label{sec:sim_data}
In the following section, we compare the performance of AEP$n$P and EP$n$P on synthetic data generated under unknown anisotropic scaling factors. The conclusions drawn from this analysis can be generalized to real data.

\paragraph{\textbf{Data preparation}}
We consider a virtual camera here with an image resolution of $640\times480$, a principal point at ($u_c$, $v_c$) = (320, 240), and a focal length of $f_u$ = $f_v$ = 150. The 3D points are randomly sampled from the interval of $[-1, 1)\times[-1, 1)\times[-1, 1)$, and the 2D points are consequently computed using the pre-defined intrinsic parameters and a random rigid transformation generated each time while ensuring that all points lie in front of the camera. Note that we randomly select independent scaling factors within the range of $[0.5, 2.0]$ to respectively scale the coordinates along the y-axis and z-axis, resulting in a new set of 3D coordinates that forms the actual synthetic 2D-3D correspondences we use in our test. All simulations are independently run 2000 times to ensure data diversity and each time we generate 1024 correspondences as the default, unless otherwise specified.

% 2000iter, 1024p
\begin{figure}[!h]
  \centering
   \includegraphics[width=\linewidth]{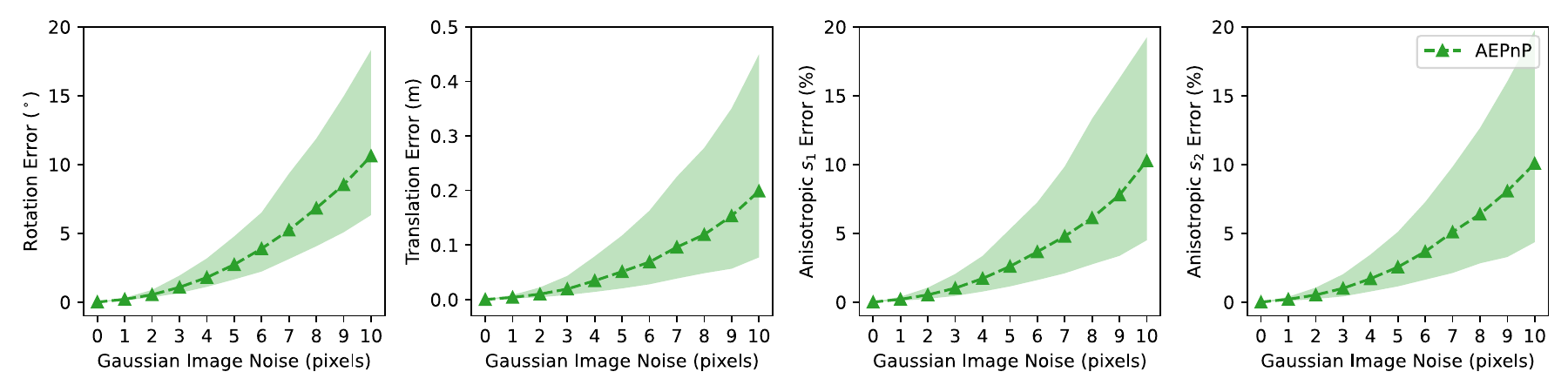}
   \includegraphics[width=\linewidth]{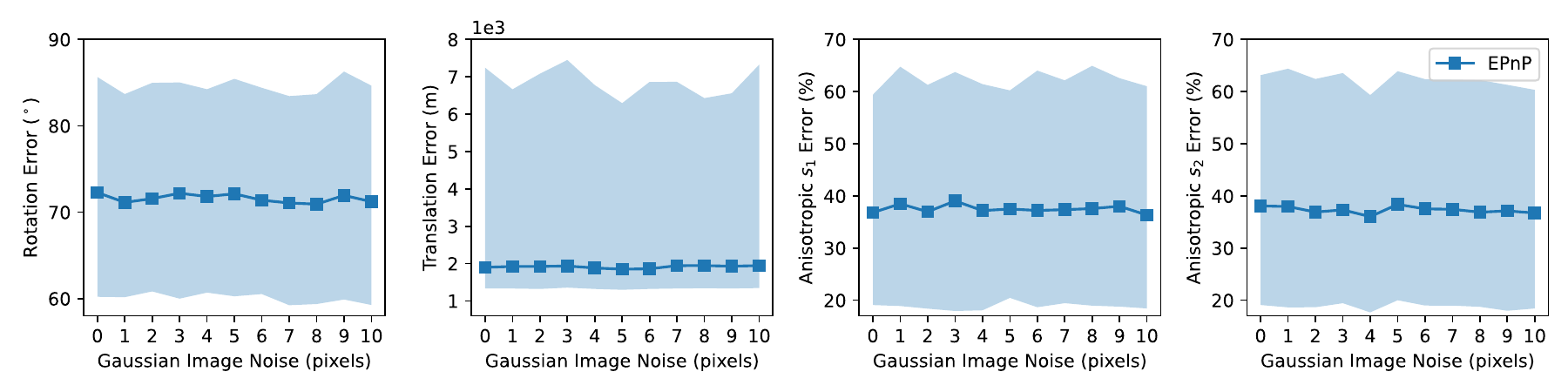}
   \caption{Statistical plot of different metrics with respect to Gaussian image noise (\#corrspondences=1024). Lines with markers represent the median errors of different methods while the corresponding shaded area indicates IQR, the difference between the 75th and 25th percentiles of the data.}
   \label{Fig:noise_exp}
\end{figure}

% 2000iter, noise_std=2
\begin{figure}[!h]
  \centering
  \includegraphics[width=\linewidth]{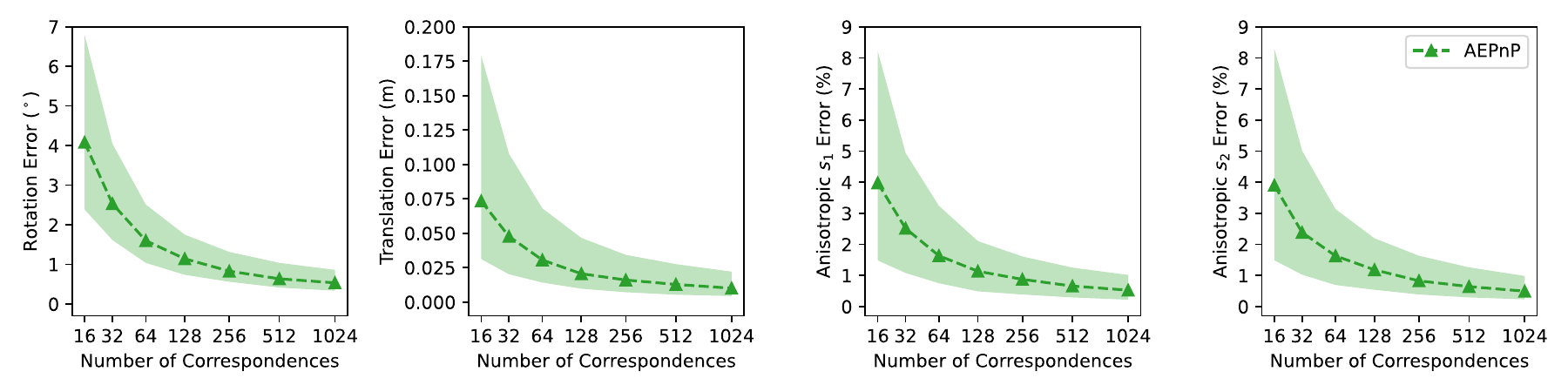}
  \includegraphics[width=\linewidth]{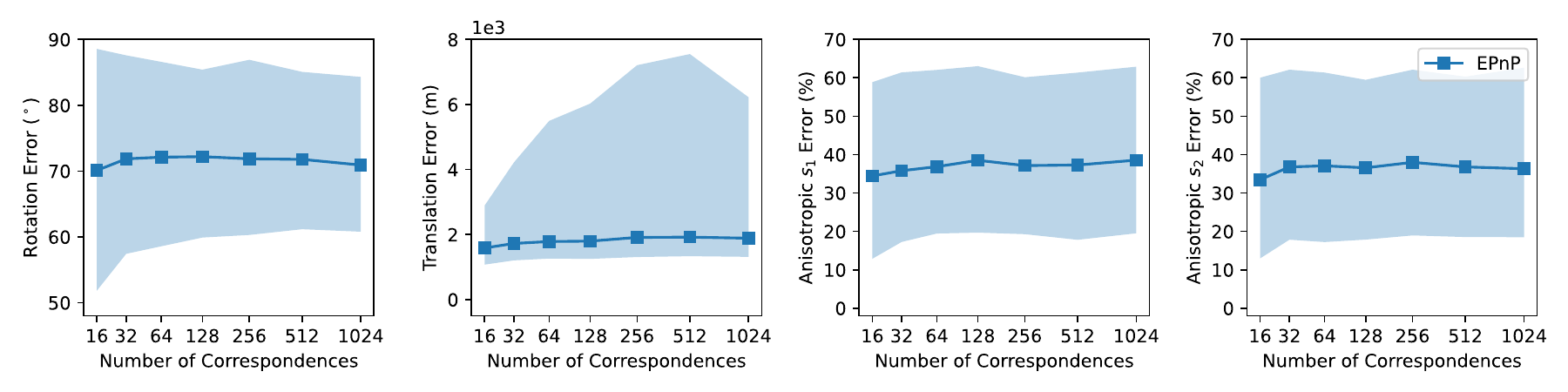}
  \caption{Statistical plot of different metrics with respect to the number of correspondences (noise=2px). Lines with markers represent the median errors of different methods while the corresponding shaded area indicates IQR, the difference between the 75th and 25th percentiles of the data.}
  \label{Fig:num_corrs_exp}
\end{figure}

\paragraph{\textbf{Robustness to noise}}
To test the robustness against noise, we add different amounts of Gaussian noise to 2D image points. The statistical results are shown in Fig.~\ref{Fig:noise_exp}. As expected, the errors for all metrics of AEP$n$P increase with the noise levels while EP$n$P completely fails even when there is no noise in the data. The interquartile range (IQR) is represented by the shaded area in the graph, indicating the stability of the solver. It is obvious that large noise can severely hinder the performance of AEP$n$P but it is still robust against certain low-level noises in input data. As for larger noise, we will show in the following that additional refinement schemes, such as RANSAC and non-linear optimization, can remarkably enhance the performance of AEP$n$P.

\paragraph{\textbf{Effect of the number of correspondences}}
We further explore the impact of the total number of correspondences used for computation by varying the number of correspondences generated in each test. Additionally, to better visualize the results, we introduce Gaussian image noise with a standard deviation of 2 pixels. As shown in Fig.~\ref{Fig:num_corrs_exp}, the errors for all metrics of AEP$n$P decrease when the total number of correspondences increases, but the rate of decrease slows down when the number of correspondences is large enough. Therefore, we suggest using as many correspondences as possible without compromising efficiency. Meanwhile, EP$n$P again fails the test no matter how many correspondences are used for computation.

\begin{figure}[!h]
  \centering
  \includegraphics[width=0.35\linewidth]{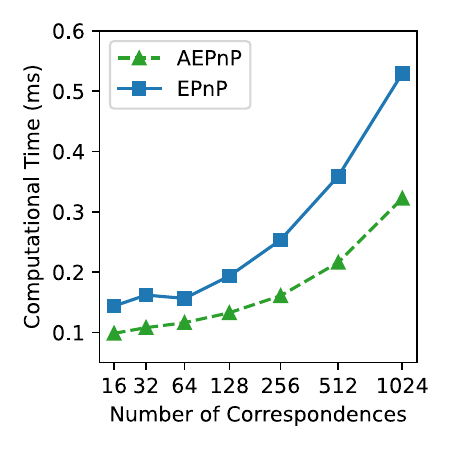}
  \caption{Computational time analysis with respect to the number of correspondences.}
  \label{Fig:time_exp}
\end{figure}

\paragraph{\textbf{Computational time analysis}}
We also investigate the efficiency of the proposed solver and report the results in Fig.~\ref{Fig:time_exp}. We calculate the average running time over 2000 independent tests with respect to different numbers of correspondences used for estimation. The computational time grows with the number of correspondences but our AEP$n$P still runs faster than EP$n$P, making it suitable for real-time applications.

% 2000iter, no noise

\subsection{Real Data}\label{sec:real_data}

In this subsection, we will mainly focus on demonstrating the effectiveness of AEP$n$P on real datasets and provide more insights regarding the integration with other refinement approaches.

\paragraph{\textbf{Datasets}}
In this section, we conduct our experiments on two real datasets, NYU-RGBD~\cite{nyu_rgbd} and MegaDepth~\cite{megadepth}. Here we use the pre-processed data provided by~\cite{liu2020learning}. Specifically, they generated 10000 2D-3D testing pairs based on the NYU-RGBD dataset, with each pair containing 1000 data points. The 3D transformations were manually generated in a certain way. The MegaDepth dataset is more challenging as it collects images from the internet and exhibits diverse camera poses. Liu Liu et al.~\cite{liu2020learning} generated 10795 2D-3D testing pairs based on the MegaDepth dataset, with the number of data points in each pair varying from tens to thousands.

We manipulate the anisotropic scaling of 3D points in the same way as for simulation data described in Sec.~\ref{sec:sim_data}. Scaling factors are randomly drawn from $[0.5, 2.0]$ to scale the coordinates along the y-axis and z-axis, generating updated 2D-3D pairs for our purpose. Note that we manually add different ratios of outlier correspondences into the input data to further analyze its robustness.

\begin{figure}[!htbp]
  \centering
   \includegraphics[width=\linewidth]{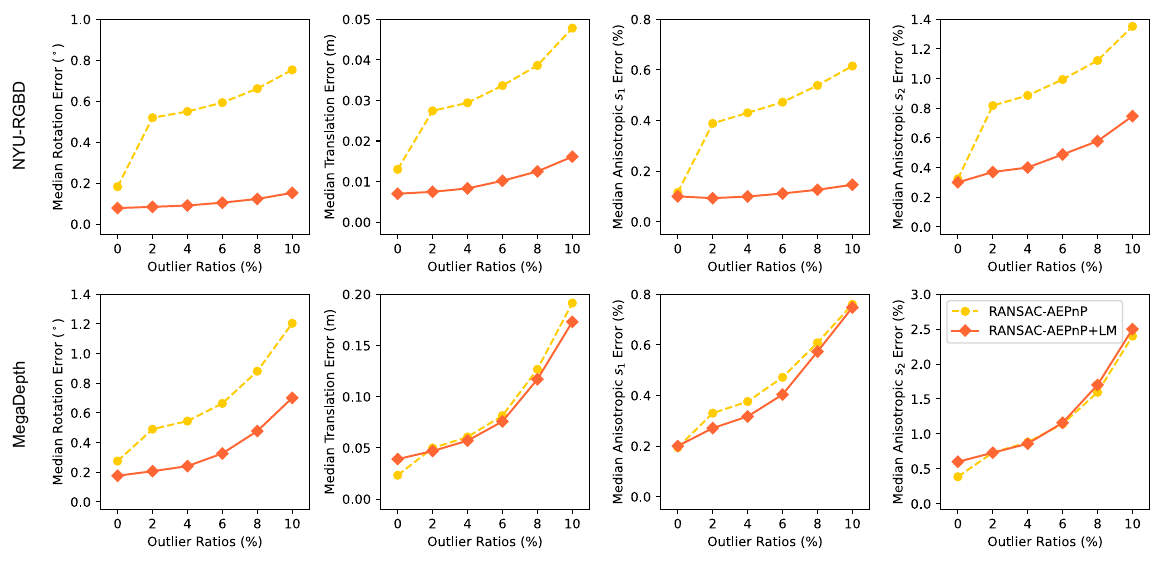}
   \caption{Different metrics with respect to outlier ratios on two real datasets.}
   \label{Fig:regular_scale_exp}
\end{figure}

\paragraph{\textbf{Results}}

In our simulation experiments in Section \ref{sec:sim_data}, we discovered that a bare AEP$n$P solver is ineffective for real-world applications due to significant noise presented in the input data. To address this, we further integrate it with a RANSAC loop and optionally apply non-linear optimization at the end to improve its ability to handle potential noise and outlier correspondences.

We evaluate the two methods on real datasets and show the results in Fig.~\ref{Fig:regular_scale_exp}. It can be seen that RANSAC-AEP$n$P successfully recovers the 6D pose as well as the two scaling factors with comparable low errors, even under a 10\% outlier ratio. With the help of the non-linear refinement step, it can achieve better results, especially on NYU-RGBD dataset.

\subsection{Sparse Keypoints Test}
We further showcase the possibility of AEP$n$P in a demanding scenario where only sparse keypoints are available for estimation. This situation frequently arises in object-related tasks, such as CAD model retrieval and alignment, where the keypoints may originate from manual annotation or keypoint detection.

\paragraph{\textbf{Dataset}}
KeypointNet \cite{you2020keypointnet} is a large-scale dataset extended from ShapeNet \cite{chang2015shapenet}, containing numerous 3D models with manually annotated 3D keypoints. We select 4 categories from KeypointNet, namely chair, bottle, car, and vessel. Each instance within the same category is rendered from a specific viewpoint to ensure the visibility of enough 3D keypoints. Finally, we obtain a subset consisting of 383 chairs, 154 bottles, 932 cars, and 312 vessels where each instance has at least 6 keypoints visible in rendered 2D images. The post-processing procedures for generating anisotropic scaled 3D coordinates are identical to those described in Sec.~\ref{sec:real_data}. To manifest the effectiveness of AEP$n$P, we add 1-pixel Gaussian noise to 2D image coordinates.

% Total chair inst. 383, Avg_num_kp 7.102:  med. rot. : 0.998, med. trans. : 0.004, med. s1 : 0.008, med. s2 : 0.012
% Total bottle inst. 154, Avg_num_kp 6.864:  med. rot. : 7.004, med. trans. : 0.100, med. s1 : 0.089, med. s2 : 0.107
% Total car inst. 932, Avg_num_kp 9.909:  med. rot. : 1.599, med. trans. : 0.011, med. s1 : 0.029, med. s2 : 0.011
% Total vessel inst. 312, Avg_num_kp 7.160:  med. rot. : 4.592, med. trans. : 0.046, med. s1 : 0.063, med. s2 : 0.041

% noise=1px
\begin{table}[!h]
\centering
\resizebox{\columnwidth}{!}{%
\begin{tabular}{c|ccccc}
\hline
\multirow{2}{*}{Category} & \multicolumn{5}{c}{Metrics}                                                                                                      \\ \cline{2-6} 
                          & \multicolumn{1}{c|}{Avg. keypoints} & Med. $R$ err. ($^\circ$) & Med. $t$ err. (m) & Med. $s_1$ err. (\%) & Med. $s_2$ err. (\%) \\ \hline
chair                     & \multicolumn{1}{c|}{7}              & 0.998                    & 0.004             & 0.008                & 0.012                \\ \hline
bottle                    & \multicolumn{1}{c|}{7}              & 7.004                    & 0.100             & 0.089                & 0.107                \\ \hline
car                       & \multicolumn{1}{c|}{10}             & 1.599                    & 0.011             & 0.029                & 0.011                \\ \hline
vessel                    & \multicolumn{1}{c|}{7}              & 4.592                    & 0.046             & 0.063                & 0.041                \\ \hline
\end{tabular}
}
\caption{Object pose estimation results of AEP$n$P on KeypointNet dataset. We manually add 1-pixel noise to manifest the effectiveness of AEP$n$P.}
\label{tab:scale}
\end{table}

% noise=1px
\begin{figure}[!h]
  \centering
   \includegraphics[width=\linewidth]{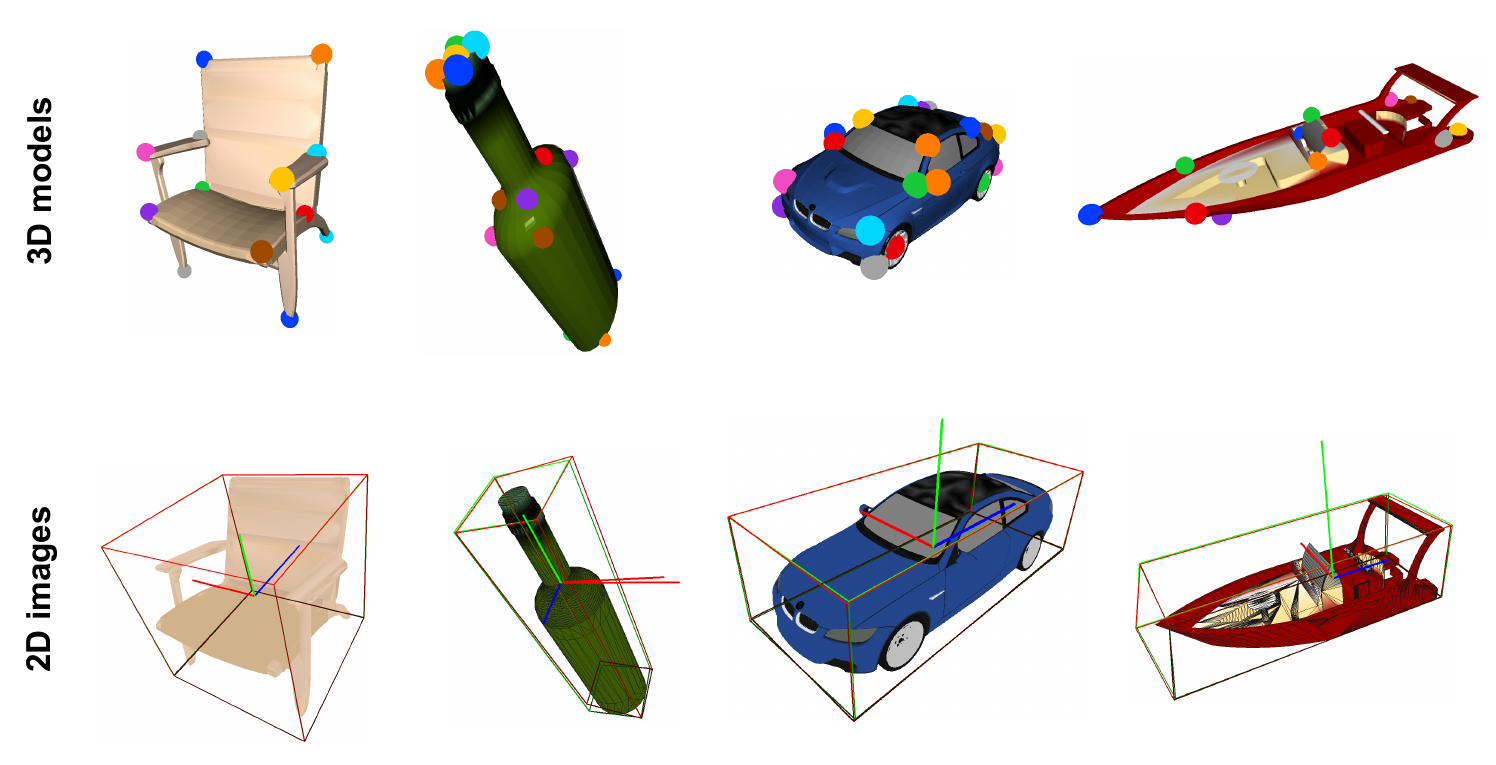}
   \caption{Qualitative results of AEP$n$P on KeypointNet dataset (noise=1px). The ground truth and predictions are denoted by green and red bounding boxes, respectively. The length of the axis also indicates the anisotropic scaling applied along that dimension.}
   \label{Fig:vis}
\end{figure}

\paragraph{\textbf{Results}}
We leverage AEP$n$P for the sparse keypoints test. The results are presented in Table~\ref{tab:scale}. Remarkably, our proposed method consistently achieves low median errors for all metrics, even with an average of 7-10 keypoint pairs. To gain a better understanding of this task, we show some qualitative results in Fig.~\ref{Fig:vis}. Although there is an anisotropic scaling between available 3D models and actual objects observed in 2D images, AEP$n$P can still predict a tight bounding box (red) surrounding the object in the 2D image, which indicates the high accuracy of our approach for solving object poses and underlying scaling factors.

\section{Conclusion}
\label{sec:conclusion}

In conclusion, we have introduced a novel 
AEP$n$P solver that reduces the heavy reliance on precise 3D models, which is typically required by traditional P$n$P solvers. This solver can estimate the 6D pose, along with anisotropic scaling factors in challenging object pose estimation tasks. Through mathematical derivations, we have revealed that this new problem shares a similar algebraic form with EP$n$P and therefore can be simply solved by a linear null-space decomposition followed by scale-aware control point registration. Extensive experiments on both simulation and real data have demonstrated the effectiveness of our proposed AEP$n$P solver, showing its potential to make pose estimation more flexible in practical settings.

% \clearpage  % TODO REVIEW/FINAL: This \clearpage needs to be removed from both review and camera-ready versions.

% ---- Bibliography ----
%
% BibTeX users should specify bibliography style 'splncs04'.
% References will then be sorted and formatted in the correct style.
%
\bibliographystyle{splncs04}
\bibliography{main}

\begin{thebibliography}{10}
\providecommand{\url}[1]{\texttt{#1}}
\providecommand{\urlprefix}{URL }
\providecommand{\doi}[1]{https://doi.org/#1}

\bibitem{abidi95}
Abidi, M., Chandra, T.: A new efficient and direct solution for pose estimation using quadrangular targets: algorithm and evaluation. IEEE TPAMI  \textbf{17}(5),  534--538 (1995)

\bibitem{ameller00}
Ameller, M.A., Triggs, B., Quan, L.: Camera pose revisited--new linear algorithms. In: In Archieve ouverte HAL: Open science (2000)

\bibitem{ansar03}
Ansar, A., Daniilidis, K.: Linear pose estimation from points or lines. IEEE TPAMI  \textbf{25}(5),  578--589 (2003)

\bibitem{banno18}
Banno, A.: A p3p problem solver representing all parameters as a linear combination  \textbf{70},  55--62 (2018)

\bibitem{bujnak08}
Bujnak, M., Kukelova, Z., Pajdla, T.: A general solution to the p4p problem for camera with unknown focal length. In: CVPR (2008)

\bibitem{chang2015shapenet}
Chang, A.X., Funkhouser, T., Guibas, L., Hanrahan, P., Huang, Q., Li, Z., Savarese, S., Savva, M., Song, S., Su, H., et~al.: Shapenet: An information-rich 3d model repository. arXiv preprint arXiv:1512.03012  (2015)

\bibitem{dementhon92}
De~Menthon, D., Davis, L.: Exact and approximate solutions of the perspective-three-point problem. IEEE TPAMI  \textbf{14}(11),  1100--1105 (1992)

\bibitem{dementhon95}
De~Menthon, D., Davis, L.: Model-based object pose in 25 lines of code. IJCV  \textbf{15}(1--2),  123--141 (1995)

\bibitem{dhome89}
Dhome, M., Richetin, M., Lapreste, J.: Determination of the attitude of 3{D} objects from a single perspective view. IEEE TPAMI  \textbf{11}(12),  1265--1278 (1989)

\bibitem{ding23}
Ding, Y., Yang, J., Larsson, V., Olsson, C., \r{A}str\"om, K.: Revisiting the p3p problem. In: CVPR (2023)

\bibitem{finsterwalder37}
Finsterwalder, S., Scheufele, W.: Das {R}\"uckw\"artseinschneiden im {R}aum. Verlag Herbert Wichmann, Berlin, Germany (1937)

\bibitem{fiore01}
Fiore, P.: Efficient linear solution of exterior orientation. IEEE TPAMI  \textbf{23}(2),  140--148 (2001)

\bibitem{fischler81}
Fischler, M., Bolles, R.: Random sample consensus: a paradigm for model fitting with applications to image analysis and automated cartography. Communications of the ACM  \textbf{24}(6),  381--395 (1981)

\bibitem{gao03}
Gao, X., Hou, X., Tang, J., Cheng, H.: Complete solution classification for the perspective-three-point problem. IEEE TPAMI  \textbf{25}(8),  930--943 (2003)

\bibitem{grafarend89}
Grafarend, E., Lohse, P., Schaffrin, B.: Dreidimensionaler {R}\"uckw\"artsschnitt, {T}eil 1: Die projektiven {G}leichungen. Zeitschrift f\"ur {V}ermessungswesen, {G}eod\"atisches {I}nstitut, {U}niversit\"at {S}tuttgart pp. 1--37 (1989)

\bibitem{grunert1841}
Grunert, J.: Das pothenotische {P}roblem in erweiterter {G}estalt nebst \"uber seine {A}nwendungen in {G}eod\"asie. In: Grunerts {A}rchiv f\"ur {M}athematik und {P}hysik (1841)

\bibitem{guo2013novel}
Guo, Y.: A novel solution to the p4p problem for an uncalibrated camera. Journal of mathematical imaging and vision  \textbf{45},  186--198 (2013)

\bibitem{haralick91}
Haralick, R., Lee, C., Ottenberg, K., Nolle, M.: Analysis and solutions of the three point perspective pose estimation problem. In: CVPR. Maui, USA (1991)

\bibitem{hesch11}
Hesch, J., Roumeliotis, S.: {A Direct Least-Squares (DLS) Method for PnP}. In: ICCV (2011)

\bibitem{horaud89}
Horaud, R., Conio, B., Leboulleux, O.: An analytic solution for the perspective 4-point problem. Computer Vision, Graphics, and Image Processing  \textbf{47}(1),  33--44 (1989)

\bibitem{hung86}
Hung, Y., Yeh, P., Harwood, D.: Passive ranging to known planar point sets. San Francisco, CA, USA (1986)

\bibitem{kanaeva15}
Kanaeva, E., Gurevich, L., Vakhitov, A.: Camera pose and focal length estimation using regularized distance constraints. In: BMVC (2015)

\bibitem{ke17}
Ke, T., Roumeliotis, S.I.: An efficient algebraic solution to the perspective-three-point problem. In: CVPR (2017)

\bibitem{kneip13}
Kneip, L., Scaramuzza, D., Siegwart, R.: A novel parametrization of the perspective-three-point problem for a direct computation of absolute camera position and orientation. In: CVPR (2013)

\bibitem{kneip2014upnp}
Kneip, L., Li, H., Seo, Y.: Upnp: An optimal o (n) solution to the absolute pose problem with universal applicability. In: ECCV. pp. 127--142. Springer (2014)

\bibitem{larsson2018camera}
Larsson, V., Kukelova, Z., Zheng, Y.: Camera pose estimation with unknown principal point. In: Proceedings of the IEEE Conference on Computer Vision and Pattern Recognition. pp. 2984--2992 (2018)

\bibitem{lepetit09}
Lepetit, V., Moreno-Noguer, F., Fua, P.: {E}{P}n{P}: An accurate {O}(n) solution to the {P}n{P} problem. IJCV  \textbf{81}(2),  578--589 (2009)

\bibitem{megadepth}
Li, Z., Snavely, N.: Megadepth: Learning single-view depth prediction from internet photos. In: Proceedings of the IEEE conference on computer vision and pattern recognition. pp. 2041--2050 (2018)

\bibitem{linnainmaa88}
Linnainmaa, S., Harwood, D., Davis, L.: Pose estimation of a three-dimensional object using triangle pairs. IEEE TPAMI  \textbf{10}(5),  634--647 (1988)

\bibitem{liu2020learning}
Liu, L., Campbell, D., Li, H., Zhou, D., Song, X., Yang, R.: Learning 2d-3d correspondences to solve the blind perspective-n-point problem. arXiv preprint arXiv:2003.06752  (2020)

\bibitem{merrit49}
Merritt, E.: Explicit three-point resection in space. Photogrammetric Engineering  \textbf{15}(4),  649--655 (1949)

\bibitem{nakano19}
Nakano, G.: A simple direct solution to the perspective-three-point problem. In: BMVC (2019)

\bibitem{penate13}
Penate-Sanchez, A., Andrade-Cetto, J., Moreno-Noguer, F.: Exhaustive linearization for robust camera pose and focal length estimation. IEEE TPAMI  \textbf{35},  2387--2400 (2013)

\bibitem{persson18}
Persson, M., Nordberg, K.: {Lambda Twist: An Accurate Fast Robust Perspective Three Point (P3P) Solver}. In: ECCV (2018)

\bibitem{quan99}
Quan, L., Lan, Z.: Linear n-point camera pose determination. IEEE TPAMI  \textbf{21}(8),  774--780 (1999)

\bibitem{schweighofer06}
Schweighofer, G., Pinz, A.: Robust pose estimation from a planar target. IEEE TPAMI  \textbf{28}(12),  2024--2030 (2006)

\bibitem{nyu_rgbd}
Silberman, N., Hoiem, D., Kohli, P., Fergus, R.: Indoor segmentation and support inference from rgbd images. In: Computer Vision--ECCV 2012: 12th European Conference on Computer Vision, Florence, Italy, October 7-13, 2012, Proceedings, Part V 12. pp. 746--760. Springer (2012)

\bibitem{sutherland63}
Sutherland, I.: Sketchpad: A man machine graphical communications system (1963), technical Report 296, MIT Lincoln Laboratories

\bibitem{triggs99}
Triggs, B.: Camera pose and calibration from 4 or 5 known 3{D} points. In: ICCV (1999)

\bibitem{wu15}
Wu, C.: {P3.5P}: Pose estimation with unknown focal length. In: CVPR (2015)

\bibitem{you2020keypointnet}
You, Y., Lou, Y., Li, C., Cheng, Z., Li, L., Ma, L., Lu, C., Wang, W.: Keypointnet: A large-scale 3d keypoint dataset aggregated from numerous human annotations. In: Proceedings of the IEEE/CVF Conference on Computer Vision and Pattern Recognition. pp. 13647--13656 (2020)

\bibitem{zheng16}
Zheng, Y., Kneip, L.: A direct least-squares solution to the {PnP} problem with unkown focal length. In: CVPR (2016)

\bibitem{zheng13}
Zheng, Y., Kuang, Y., Sugimoto, S., Astrom, K., Okutomi, M.: Revisiting the {P}n{P} problem: A fast, general and optimal solution. In: ICCV (2013)

\bibitem{zheng14}
Zheng, Y., Sugimoto, S., Sato, I., Okutomi, M.: A general and simple method for camera pose and focal length determination. In: CVPR (2014)

\end{thebibliography}
\end{document}